\documentclass[a4paper,oneside,11pt]{article}

\usepackage[american]{babel}
\usepackage[utf8]{inputenc}

\usepackage{csquotes}

\usepackage{subcaption}
\usepackage{calc}

\usepackage{amssymb}
\usepackage{amstext}

\usepackage{pslatex}
\usepackage[bottom]{footmisc}

\usepackage{setspace}
\usepackage{xspace}
\usepackage{xurl}
\usepackage{authblk}

\usepackage{natbib}

\usepackage{xcolor}

\usepackage[acronym, nomain, section]{glossaries}
\setacronymstyle{long-short}

\usepackage{tabularx}
\usepackage{booktabs}
\usepackage{makecell}

\usepackage{array}
\newcolumntype{L}{>{\raggedright\arraybackslash}X}
\newcolumntype{C}{>{\centering\arraybackslash}X}

\usepackage{csvsimple}

\usepackage{microtype}
\usepackage{needspace}

\usepackage[hidelinks]{hyperref}
\usepackage{cleveref}
\usepackage{orcidlink}

\renewcommand{\citep}[1]{\cite{#1}}
\renewcommand{\citet}[1]{\cite{#1}}

\newcommand{\OCCUPATION}{\mbox{\textsc{\small OCCUPATION}}\@\xspace}
\newcommand{\BOCCUPATION}{\mbox{\textsc{\small B-OCCUPATION}}\@\xspace}
\newcommand{\IOCCUPATION}{\mbox{\textsc{\small I-OCCUPATION}}\@\xspace}

\newcommand{\eg}{e.\,g.\@\xspace}
\newcommand{\ie}{i.\,e.\@\xspace}

\renewcommand{\Pr}{\text{Pr}} 
\renewcommand{\Re}{\text{Re}} 
\newcommand{\FOne}{$\text{F}_1$\@\xspace} 
\newcommand{\TP}{\text{TP}}
\newcommand{\FP}{\text{FP}}
\newcommand{\FN}{\text{FN}}

\newcommand{\conf}{\text{conf}}

\newacronym{CRF}{CRF}{Conditional Random Field}
\newacronym{CVET}{CVET}{Continuing Vocational Education and Training}

\newacronym{DAPT}{DAPT}{Domain-adaptive Pre-training}

\newacronym{CNN}{CNN}{Convolutional Neural Network}

\newacronym{DNN}{DNN}{Deep Neural Network}

\newacronym{FRG}{FRG}{Federal Republic of Germany}

\newacronym{HMM}{HMM}{Hidden Markov Model}

\newacronym{IE}{IE}{Information Extraction}

\newacronym{GDR}{GDR}{German Democratic Republic}

\newacronym{LLM}{LLM}{Large Language Model}
\newacronym{LSTM}{LSTM}{Long short-term memory}

\newacronym{NAT}{NAT}{Noise-aware Training}
\newacronym{NER}{NER}{Named Entity Recognition}
\newacronym{NLP}{NLP}{Natural Language Processing}

\newacronym{OCR}{OCR}{Optical Character Recognition}
\newacronym{OJA}{OJA}{Online Job Advertisement}

\newacronym{SVM}{SVM}{Support Vector Machine}

\newacronym{VET}{VET}{Vocational Education and Training}
\newacronym{VLM}{VLM}{Vision Language Model}

\graphicspath{
    {figures/}
}

\title{Two-Step Occupation Coding}

\author[1]{Alexander M. Esser\,\orcidlink{0000-0002-5974-2637}}
\author[1,2,3]{Jens Dörpinghaus\,\orcidlink{0000-0003-0245-7752}}

\affil[1]{University of Koblenz, Germany}
\affil[2]{Federal Institute for Vocational Education and Training (BIBB), Bonn, Germany}
\affil[3]{Linnaeus University (LNU), Växjö, Sweden}

\affil[ ]{\texttt{alexanderesser@uni-koblenz.de}}

\date{}

\begin{document}

\maketitle

\begin{center}
    \small
    \noindent Preprint of the paper accepted for the \\Federated Conference on Computer Science and Information Systems\\ (FedCSIS 2026)
\end{center}

\vspace{2em}


\begin{abstract}
    Occupation coding links job titles in free text to occupational taxonomies and is a core task in labor market research. Existing approaches typically address this problem in a single end-to-end step, jointly identifying job titles and assigning occupational codes. This paper presents a novel two-step approach that separates these tasks. In the first step, a domain-specific Named Entity Recognition (NER) model identifies occupational titles in continuous text, even under noise such as OCR errors. In the second step, the extracted job titles are mapped to a taxonomy, enabling the classifier to focus exclusively on this mapping.
    We demonstrate that this separation improves accuracy, robustness, and interpretability compared to single-step approaches.
    The method has been developed for German documents but is transferable to other languages.
    We further introduce a margin-based confidence criterion for occupation coding, replacing common absolute thresholds.
    To support reproducibility, we publish the source code and evaluation scripts.
\end{abstract}

\vfill

{
\footnotesize
\noindent \copyright~2026 IEEE. Personal use of this material is permitted.
Permission from IEEE must be obtained for all other uses, including
reprinting/republishing this material for advertising or promotional
purposes, collecting new collected works for resale or redistribution
to servers or lists, or reuse of any copyrighted component of this
work in other works.
}

\pagebreak[3]


\section{Introduction}
\label{sec:introduction}

Occupation coding (OC) is the process of linking a given job title to a predefined occupational taxonomy. It can be understood as a text classification problem in which a textual sequence is assigned to one of potentially thousands of classes.
Automatically classifying job titles plays a central role in labor market research \citep{Dorau2025b,Gweon2017, Schierholz2014,Schierholz2021}. Job titles are important for job seekers when searching for relevant job postings, for HR departments when filling new positions, and for vocational institutions when aligning their programs with labor market demands \citep{Decorte2021}. Although job titles generally follow a certain structure, they are often freely entered as unstructured text \citep{VanHautte2020}.

Usually, for occupation coding, end-to-end approaches are used, which simultaneously identify job titles and map them to standardized occupational taxonomies. Such single-step approaches, however, mix two conceptually distinct tasks: (i)~identify whether a given text span is a job title at all, and (ii)~classify a recognized title according to a taxonomy.
We propose a two-step occupation coding framework that deliberately separates these tasks.

In the first step, a domain-specific NER model identifies job titles in continuous text, ensuring that only legitimate job title mentions are passed to the subsequent classification stage.
This step builds upon our previous work \citep{Esser2026} on \gls{NER} in the field of \gls{VET}, which provides an NER model for detecting occupational titles and other entity types in noisy textual contexts.
In the second step, the identified job titles are mapped to a taxonomy. This division of tasks yields a noise-reduced input for occupation coding, allowing the classifier to focus on the title-taxonomy alignment rather than first filtering out non-occupational phrases. From a conceptual point of view, this improves accuracy by reducing the risk of false positive matches.

\begin{figure}[tbp]
    \centering
    \includegraphics[width=0.9\columnwidth]{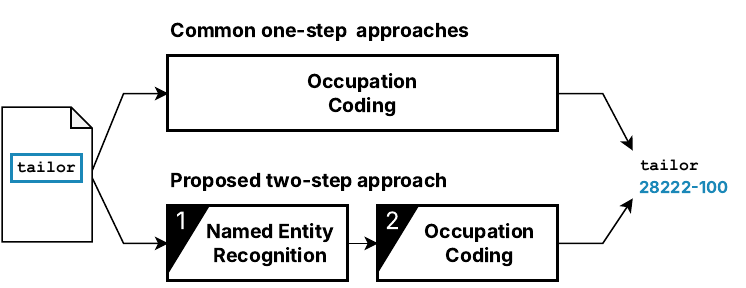}
    \caption{One-step and two-step occupation coding approaches.}
    \label{fig:two_step}
    \vspace{0em plus 1em}
\end{figure}

\Cref{fig:two_step} illustrates the core idea: Within a VET document, the job title \enquote{tailor} is recognized. The corresponding job ID is retrieved from a taxonomy. While common approaches perform identification and classification as one single end-to-end task, the proposed method follows a two-step strategy, first applying \gls{NER}, and then matching all recognized job titles against a taxonomy.
The proposed methods, in particular the components for \gls{NER} and occupation coding, have been developed for German documents but are transferable to other languages.

In Germany, the reference taxonomy for occupational titles is the \emph{German Classification System of Occupations} (Klassifikation der Berufe; KldB), used by the Federal Employment Agency (BA) and its research institute (IAB).
Here, occupations are structured at task level. The latest version is the 2020 revision of KldB 2010. It has been developed for compatibility with the \emph{International Standard Classification of Occupations\footnote{See \url{https://www.ilo.org/public/english/bureau/stat/isco/isco08/}.}} (ISCO) \citep{Fischer2024,Doerpinghaus2023}.
ISCO occupations are structured by skill level and linked to the \emph{European Skills, Competences, Qualifications and Occupations} (ESCO) ontology~\citep{Reiser2024, Dorau2025}.

\pagebreak[1]

The German Classification System of Occupations KldB \citep{KldB2020} defines a five-level hierarchy of groups, represented by five-digit codes. The first digit denotes the general occupational area (\emph{Berufsbereich}), down to the fifth digit, which specifies the detailed occupational activity (\emph{Berufsgattung}).
Each job title is allocated to the lowest of these five groups, \eg, the profession of \emph{pharmaceutical technical assistant} (\emph{Pharmazeutisch-technischer Assistent}; code 81822-105) is assigned to group 81822.
Occupation coding approaches typically predict the 5-digit codes, \ie, the occupational group at the lowest level.

While pipeline decomposition is common in information extraction, this paper is one of the first to apply this idea to occupation coding, presenting a two-step approach.
Our method demonstrates that splitting occupation coding into two distinct tasks -- separately identifying and matching job titles -- measurably improves accuracy, robustness, and interpretability.


The main contributions of this paper are:

\begin{itemize}
    \item We propose a novel \emph{two-step occupation coding approach} that explicitly separates job title identification from taxonomy-based occupation classification, in contrast to existing single-step, end-to-end approaches.

    \item For \emph{NER}, we developed a domain-specific model, which is trained using a noise-aware, multi-step fine-tuning strategy integrating occupational taxonomies as additional training resources.

    \item For \emph{occupation coding}, we retrain the model to focus exclusively on matching previously identified job titles against the taxonomy. We systematically compare multiple classifiers, including Support Vector Machines, logistic regression, and XGBoost. We demonstrate that a two-step approach substantially improves accuracy, robustness, and interpretability.

    \item Furthermore, we introduce a \emph{margin-based confidence criterion} for occupation coding as an alternative to commonly used absolute confidence thresholds.
\end{itemize}

To support the reproducibility of scientific results, we publicly provide the source code and evaluation scripts for our two-step occupation coding approach.\footnote{See \url{https://github.com/TM4VETR/two_step_occupation_coding}.}


The remainder is structured as follows:
\Cref{sec:related_work} discusses relevant literature in the fields of \gls{NER} and Occupation Coding.
\Cref{sec:method} presents the proposed approach.
\Cref{sec:evaluation} evaluates the results.
\Cref{sec:conclusion} addresses limitations and directions for future work, and concludes this study.

\pagebreak[3]


\section{Related Work}
\label{sec:related_work}

In this section, existing approaches in the fields of \gls{NER} and occupation coding are reviewed.

\subsection{Named Entity Recognition}
\label{sec:related_work_ner}

The aim of \gls{NER} is to identify named entities in continuous text according to a set of predefined categories \citep{Nadeau2007}.
It can be regarded as a particular instance of the sequence labeling problem, which concerns assigning labels to tokens in an input sequence~\citep{TjongKimSang2003}.
A common distinction can be made between classical \emph{heuristic-based} and modern \emph{deep learning-based} \gls{NER} approaches.

\pagebreak[1]

\paragraph{Heuristic-based Approaches}
Within heuristic-based approaches, a further distinction can be made between \emph{rule-based} and \emph{statistical} methods:
Rule-based approaches, such as pattern matching techniques and regular expressions in particular, rely on manually crafted rules to detect entities \citep{Eftimov2017}.
These methods typically depend on curated lists of known entities.
Statistical methods, including \glspl{HMM} \citep{Bikel1999} and \glspl{CRF} \citep{McCallum2003}, formulate \gls{NER} as a sequence labeling task using hand-engineered features \citep{Jurafsky2025}.

\glsreset{HMM}
\glsreset{CRF}
\glspl{HMM} model sequences by transition matrices, which represent the joint probability of transitions between hidden input states and observable output states \citep{Rabiner1989}.
\glspl{CRF} are undirected graphical models that estimate the conditional probability of a sequence of labels given a corresponding input sequence \citep{Lafferty2001}.
For both \glspl{HMM} and \glspl{CRF}, algorithms such as the Viterbi algorithm~\citep{Viterbi1967} are commonly employed to infer the most probable sequence of states or labels.

\paragraph{Deep Learning-based Approaches}
\glsreset{DNN}
In contrast, deep learning-based methods apply modern learning techniques based on \glspl{DNN}.
These approaches can be further categorized according to the underlying architecture:
\emph{LSTM-based models}, in particular bi-directional \glsplural{LSTM}, are frequently used to model contextual information and are often combined with \glspl{CRF} for structured sequence decoding \citep{Huang2015, Lample2016}.
Such models typically rely on word- or character-level embeddings, including GloVe, FastText, or FLAIR~\citep{Pennington2014, Bojanowski2017, Grave2018, Akbik2018}.

\emph{Transformer-based models}, such as BERT and its numerous variants, currently define the state of the art in \gls{NER} \citep{Devlin2019}. Transformer architectures are composed of stacked self-attention layers that capture pairwise dependencies between tokens. This enables each layer to access the full input sequence while remaining parallelizable during training \citep{Vaswani2017}.
These models do not require explicit dictionaries of known entities, but instead rely on contextual representations learned from large corpora.

\subsubsection{Application to VET}
A limited number of publications have addressed \gls{NER} specifically for job titles in VET data.
\citet{Reiser2025} detected and classified occupations in German texts, comparing a rule-based approach with a language model-based approach.
\citet{Safikhani2023} fine-tuned BERT and \mbox{GPT-3} for the automated extraction of German occupations.
\citet{Decorte2021} proposed JobBERT, a deep learning-based model for recognizing job titles and linking them to ESCO classes. Starting from BERT as a pre-trained model, the recognition is based on skill information, showing that skill descriptions are an essential component for job title recognition.

\pagebreak[2]

\subsection{Occupation Coding}
\label{sec:related_work_occupation_coding}

Occupation coding describes the task of mapping a given job title to a predefined occupational taxonomy, such as the German KldB. It can be viewed as a text classification problem in which a textual input is mapped to one of potentially hundreds of possible classes \citep{Gweon2017, Schierholz2014, Schierholz2021}.

\pagebreak[1]

Automatically classifying job titles plays a central role in labor market research, for instance in the analysis of surveys or administrative data \citep{Dorau2025b}.

\pagebreak[1]

Research on occupation coding has a long tradition and dates back to a time before the era of computers \citep{National1980}.
For many years, occupational assignments were performed manually. Automation promised lower processing costs, faster production of statistics, and higher consistency in coding decisions \citep{Schierholz2021, Dorau2025b}.

In the field of occupation coding, the work of Malte Schierholz is noteworthy. His contributions include a Master's thesis \citep{Schierholz2014}, a PhD thesis \citep{Schierholz2019}, and numerous further publications \citep{Schierholz2021, Schierholz2018, Schierholz2018b, Bethmann2014}.
Starting from an early prototype for coding during interviews, these works iteratively improved the methodology and introduced new approaches.
\citet{Schierholz2021} also developed the \emph{occupationCoding} R package\footnote{See \url{https://github.com/malsch/occupationCoding}.}.
This package has been used as a reference for one-step approaches. For integration into Python-based code, we developed a Python wrapper\footnote{See \url{https://github.com/TM4VETR/occupation_coding_python}.}.

Analogous to \gls{NER}, occupation coding approaches can be grouped into \emph{rule-based methods} and \emph{learning-based methods}.

Rule-based methods rely on predefined dictionaries of occupational titles and apply pattern matching or handcrafted rules to assign codes.
While these methods are transparent and often work well with standardized terminology, they tend to perform poorly on noisy text (\eg, OCR errors or misspellings), creative job titles (\eg, in \glspl{OJA}), or historical language.
Moreover, they require ongoing manual maintenance, as labor market terminology changes over time.

Learning-based approaches apply statistical or machine learning techniques for classification.
\glsreset{SVM}
Classical techniques include \glspl{SVM}, $k$-nearest neighbors, logistic regression, naive Bayes, and random forests \citep{Creecy1992, Russ2016, Takahashi2014, Gweon2017, Ikudo2019}.
A comparative evaluation of several of these models was provided by \citet{Dorau2025b}.
More recently, deep learning methods -- typically based on Transformer architectures or \glspl{CNN} -- have been applied for occupation coding.
In \citep{Javed2015}, \emph{Carotene} was proposed, a cascade of an \gls{SVM} and $k$-nearest neighbor.
In \citep{Wang2019}, a \gls{CNN}, \emph{DeepCarotene}, was applied for classifying job titles in online advertisements.

While all of the above methods regarded occupation coding as an end-to-end approach, our approach considers occupation coding as two conceptually distinct tasks.


\section{Proposed Method}
\label{sec:method}

In this section, the proposed two-step occupation coding approach is presented. In the first step, a domain-specific \gls{NER} model is applied. In the second step, the extracted job titles are mapped to the KldB taxonomy, enabling the classifier to focus exclusively on the title-taxonomy matching.


\subsection{Named Entity Recognition}
\label{sec:method_ner}

One key challenge for \gls{NER} is noisy input data. Noise introduced upstream, such as \gls{OCR} errors or spelling mistakes, can substantially affect \gls{NER}.
Therefore, we applied a \gls{NAT} approach that improves the robustness of \gls{NER} systems in the presence of \gls{OCR} and spelling errors.

\glsreset{FRG}
\glsreset{GDR}
A second challenge arises from the variability of job titles, as they evolve over time. The historical \gls{VET} documents originate from different decades and from two German states, the \gls{FRG} and the \gls{GDR}, with different socio-political systems.
Additional variations arise from language-specific characteristics.
\citet{Reiser2025} discussed the unique challenges in the German language. One source of variation is gender-specific forms of job titles.

Given these challenges, rule-based approaches are not promising. Instead, a Transformer-based model has been trained, applying transfer learning, \gls{NAT} and multi-stage fine-tuning, specifically adapted for job titles in German \gls{VET} data.

The entity \OCCUPATION refers to either a specific job title or a higher-level group of occupations.
In previous work \citep{Esser2026, Esser2026a}, a finer distinction was made between \emph{job titles} and higher-level \emph{job title groups}. However, confusion analysis revealed difficulties in distinguishing between these entities. Ideally, entity classes should be distinct; therefore, both were merged into the single entity class \OCCUPATION.
Very general terms (\eg, the term \enquote{occupation} itself) should not be recognized as \OCCUPATION.


\subsubsection{Training Data}
\label{sec:method_ner_data}

The training data consist of \emph{historical} and \emph{contemporary} data. These data sources differ in origin, noise characteristics, and their role in the proposed two-step approach.

The historical data consist of VET documents from East and West Germany as well as reunified Germany, especially training regulations and systematic listings of job titles. They cover the period since 1976 and comprise 159 document pages in total. Since these documents were digitized from historical sources, they may contain OCR errors.

The contemporary data were crawled from X (formerly Twitter) \citep{Hein2025, Tiemann2025}. This dataset is born-digital and, therefore, does not contain OCR errors. However, spelling variations and errors may occur. Moreover, job titles frequently appear in non-standardized forms, posing challenges for both entity identification and occupation coding.

The two data sources play different roles in the proposed approach. For NER, both the historical and contemporary data were used as training data. For occupation coding, only the contemporary data were used, since this dataset contains occupations annotated according to the KldB taxonomy. Thus, both datasets contribute to entity identification, while the KldB-annotated contemporary data provide the basis for the subsequent occupation-classification step.

\begin{table}[t]
    \caption{Amount of annotated data used for NER and OC.}
    \label{tab:data}
    \centering
    \begin{tabularx}{\columnwidth}{CCC}
        \toprule
                \multicolumn{2}{c}{\textbf{NER}} & \textbf{OC}\\
tokens & entities & occupations\\
        \midrule

        \begin{filecontents*}{data/annotations.csv}
tokens;entities;jobids
62\,379\,520;1\,761\,261;1\,641\,486
\end{filecontents*}
\csvreader[
        head to column names,
        separator=semicolon,
        late after line=\\
        ]{data/annotations.csv}
        {}
        {\tokens & \entities & \jobids}

        \bottomrule
    \end{tabularx}
\end{table}

\Cref{tab:data} reports the amount of annotated data used for NER (final model within the multi-stage pipeline) and for OC. The NER data are counted at token level, including \enquote{O} tokens, while the OC data are counted at the level of occupational entities.

\pagebreak[2]


\subsubsection{Additional Training Data}
\label{sec:method_ner_additional_data}

To increase the accuracy, additional training data have been integrated. These include word lists of search terms provided by the Federal Employment Agency\footnote{\textls[-40]{See\,\url{https://www.arbeitsagentur.de/institutionen/dkz-downloadportal}.}} as well as the KldB taxonomy \citep{KldB2020, KldB1988} with its hierarchical structure. For the KldB, both the 2020 and the 1988 versions were integrated in order to account for both contemporary and historical occupational titles.

However, it is important to note that all these additional data differ from real-world data in two important respects:
Every word belongs to an entity; the data contain no \enquote{O} labels stating that the corresponding token is not part of any entity. Besides, the data contain no \gls{OCR} noise.

Therefore, the additional training data cannot be used directly for the final model. Instead, we applied multi-stage fine-tuning, introducing an intermediate pre-training stage trained on this data, as described later.

\subsubsection{Transfer Learning}
\label{sec:method_ner_transfer}

Transfer learning is a common concept in which pre-trained models, which have been trained on large datasets, are fine-tuned for target tasks on smaller, domain-specific datasets.
\citet{Namysl2019} demonstrated that a backbone model (in their case, for \gls{OCR}) can be significantly improved by adding only a small amount of data.

As base model, we use a German, cased BERT variant\footnote{See\,{}\url{https://huggingface.co/dbmdz/bert-base-german-cased}.}, released by the Bavarian State Library, with weights initialized with the original BERT values.

\pagebreak[2]

\subsubsection{Noise-Aware Training}
\label{sec:method_ner_nat}

An in-depth analysis of robustness in document processing pipelines is provided by \citet{Namysl2023a}.
The study shows that \emph{domain-specific retraining}, \ie, fine-tuning an existing model on a collection of domain-specific documents, often already leads to considerable performance gains.
A more sophisticated strategy is the application of \emph{\gls{NAT}}.

\glsreset{NAT}
\gls{NAT} refers to a set of methods aiming to make downstream \gls{NLP} systems like \gls{NER} robust to noise introduced upstream, \eg, \gls{OCR} errors.
The core idea of \gls{NAT} is to inject realistic errors into the training data so that the model learns representations that are less sensitive to such perturbations \cite{Namysl2020, Xu2021}.
In practice, this involves injecting typical \gls{OCR} or spelling errors.

\gls{NAT} is closely related to \emph{Empirical Error Modeling}, which empirically analyzes typical \gls{OCR} errors, in order to subsequently inject realistic perturbations~\citep{Namysl2021}.
For the proposed method, NAT has been used as an online approach, dynamically injecting synthetic OCR errors (insertions, deletions, substitutions) on-the-fly during training, generating a new noisy variant of each input instance before tokenization.


\subsubsection{Multi-Stage Fine-Tuning}
\label{sec:method_ner_multi_stage}

To integrate the additional training data, we introduce an intermediate pre-training stage, applying \emph{\gls{DAPT}}~\citep{Gururangan2020}.

\begin{figure}[t]
    \centering
    \includegraphics[width=0.7\columnwidth]{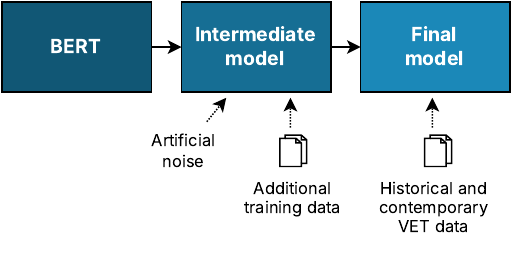}
    \caption{Multi-stage training process.}
    \label{fig:multi_stage}
    \vspace{0em plus 0.5em}
\end{figure}

\pagebreak[1]

\begin{samepage}
    \Cref{fig:multi_stage} shows the multi-stage training process, with BERT as base model, the intermediate stage pre-trained on the additional data, and the final stage trained on the real-world VET data.
\end{samepage}

\pagebreak[2]

\subsubsection{Implementation Details}
\label{sec:method_ner_implementation}

The framework for training the NER models was built on PyTorch \citep{Paszke2019} and the Hugging Face Transformers library \citep{Transformers}.

\paragraph{Bias}
In the present setting, different sources of bias may arise: pre-trained model bias, data bias, and a classification bias.
When using pre-trained models, these may be subject to bias present in their training data and affecting their predictions. This bias is carried over to the fine-tuned model~\citep{Wang2023}.
Moreover, in the data, certain occupations, time periods, or document types may be over- or underrepresented, which can influence both entity recognition and subsequent occupation classification.
Finally, the classification step depends on existing occupational taxonomies, which themselves reflect institutional classification decisions.
While a systematic bias analysis is beyond the scope of this paper, the presented model can serve as a basis for future evaluations to identify potential imbalances.

\pagebreak[2]

\paragraph{Hyperparameters}
When training the intermediate and the final model, we searched for the lowest validation loss using the following hyperparameters:
During a warm-up phase (10\,\% of the steps), the learning rate is gradually increased up to the target rate of $2\cdot10^{-5}$ (learning rate scheduling) and subsequently decreased as training progresses. Especially when fine-tuning pre-trained models, smaller learning rates allow for more refined adjustments.
The best-performing model over all epochs, measured by the $F_1$ score (micro-averaged on entity level), was saved.

\paragraph{Class Weighting and Oversampling}
To compensate for the significant imbalance, \emph{class weighting} and \emph{oversampling} have been applied. The majority of labels is \enquote{O}, stating that the corresponding token is not part of any entity. The loss function would be dominated by these labels, and the model could degenerate into predicting only \enquote{O} (class collapse). A certain proportion of \enquote{O} tokens was deliberately removed, as the model can learn little from these tokens.
The weights for class weighting have then been normalized and clamped to avoid extreme values.
Oversampling, during training, replicates token windows that contain positive labels (non-\enquote{O}) and thereby increases the relative frequency of positive examples presented to the model. This artificially rebalances the class distribution and strengthens the loss signal for rare entities.

\needspace{\baselineskip}
For pre-training the intermediate model, slightly different hyperparameters and settings have been used: The additional training data for the intermediate model contain no \enquote{O} labels. Therefore, no oversampling, weight balancing, or clamping has been applied.

\paragraph{Annotations}
For annotating entities, we used the \emph{BIO} format, in which each word (token) is assigned a tag. The tag \BOCCUPATION indicates that the token marks the beginning of an entity span, \ie of a job title or a higher-level occupational group; \IOCCUPATION indicates that the token lies inside an entity span; \enquote{O}, outside any entity span.

\pagebreak[2]

\subsection{Occupation Coding}
\label{sec:method_occupation_coding}

When applying a two-step occupation coding approach, the classifier needs to be retrained, as it no longer needs to identify job titles in free text. Instead, it can focus exclusively on matching a given job title to an occupational taxonomy, starting from a set of previously identified job titles.

\subsubsection{Classifiers}
\label{sec:method_occupation_coding_classifiers}

For retraining, we systematically compared three types of classifiers, an \gls{SVM}, XGBoost as a tree-based approach, and logistic regression.
For training and evaluating these classifiers, the \texttt{\small scikit-learn} Python package \citep{scikit-learn} was employed.
All three classifiers require input data in the form of numerical vectors. To generate these embeddings, a TF-IDF vectorizer was used.
We deliberately focused on classical classifiers to ensure comparability with the evaluated one-step occupation coding approach and to isolate the effect of task separation. Training an advanced classifier, such as a Transformer-based model, is left for future work.

From a theoretical perspective, one might assume that decision trees can best reflect the hierarchical structure of the KldB taxonomy.
However, XGBoost does not automatically exploit label hierarchies. It still treats all classes as flat categories.
The results show that the SVM achieves the best performance among the evaluated classifiers (see \Cref{sec:evaluation_results_classifiers}).

A linear SVM represents each sample as a point in a high-dimensional feature space.
It then learns linear decision boundaries (hyperplanes) that split each space into two half-spaces.
To effectively discriminate among many classes, as in the KldB taxonomy, a high number of dimensions is necessary.

In this use case, the SVM yields better performance than a tree-based method. Therefore, we selected an SVM as classifier for subsequent steps.


\subsection{Confidence Thresholding}
\label{sec:method_thresholding}

\begin{figure}[tbp]
    \centering
    \begin{subfigure}{\columnwidth}
        \centering
        \includegraphics[width=0.6\columnwidth]{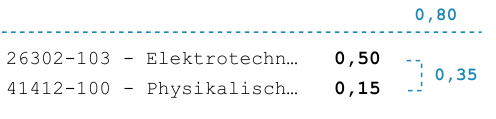}
        \caption{}
    \end{subfigure}
    \vspace*{0.5em}
    \begin{subfigure}{\columnwidth}
        \centering
        \includegraphics[width=0.6\columnwidth]{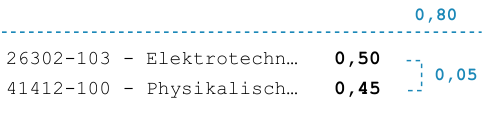}
        \caption{}
    \end{subfigure}

    \caption{Absolute threshold and confidence margin -- two examples.}
    \label{fig:confidence}
    \vspace{0em plus 0.5em}
\end{figure}

The central idea of two-step occupation coding is to allow the classifier to focus exclusively on matching a previously identified job title to a taxonomy. The confidence\footnote{The resulting probabilities are not confidences in a strict sense, but they are, after calibration, a suitable measure for comparing thresholding approaches.} associated with this decision is a critical parameter.

The combined confidence \emph{$\conf$} for the overall process is computed as the product of the NER and OC confidence:
\begin{equation}
    \conf = \conf_{\text{NER}} \cdot \conf_{\text{OC}} \, .
\end{equation}

As an additional improvement, besides the two-step approach, we introduced a confidence margin approach for occupation coding.
Common approaches, including the \emph{occupationCoding} R package \citep{Schierholz2021}, the \emph{Carotene} classifier \citep{Javed2015}, and the \emph{DeepCarotene} \gls{CNN} \citep{Wang2019}, typically rely on an absolute threshold to determine whether the algorithm is sufficiently confident about a prediction. Only if the confidence score (\ie, the predicted probability assigned by the classifier) is above this threshold is the job ID assigned to the job title.

We, in contrast, propose a margin-based approach, as commonly used in other application domains, such as computer vision. The key idea is to replace the absolute threshold with a margin defined as the difference between the highest and the second-highest predicted confidence score. Rather than asking whether a prediction exceeds a fixed threshold, the margin-based approach evaluates how confident the algorithm is about its top prediction relative to the second-highest one.

\Cref{fig:confidence} illustrates this idea in two examples. In both cases, the job ID with the highest confidence is \enquote{26302-103 -- Elektrotechnische/r Assistent/in} (\emph{electrical engineering assistant}), and the second-highest is \enquote{41412-100 -- Physikalisch-technische/r Assistent/in} (\emph{physical--technical assistant}).
In neither example does the confidence exceed the threshold of 0.8. Consequently, using an absolute threshold approach, both examples would be classified as \enquote{not recognized}.

However, in \Cref{fig:confidence}(a), there is a large difference between the two confidence scores, whereas in \Cref{fig:confidence}(b) this difference is small.
In the first example, a margin-based approach, with a suitably chosen margin, would return the top job ID. In contrast, in the second example, no job ID would be recognized, since the algorithm is not sufficiently confident in distinguishing between the two most likely candidates.
The recognition rates obtained using both approaches are reported in \Cref{sec:evaluation_results_confidence}.


\section{Evaluation}
\label{sec:evaluation}

In this section, the proposed method is evaluated, demonstrating that task separation allows error sources to be attributed more clearly to either identification or classification.

\needspace{6\baselineskip}

Standard evaluation metrics for \gls{NER} and occupation coding include \emph{precision}, \emph{recall}, and the \emph{\FOne score}.
These metrics are derived from the numbers of \emph{true positives} $\TP$, \emph{false positives} $\FP$, and \emph{false negatives} $\FN$:

\begin{samepage}
    \begin{equation}
        \begin{aligned}
            \Pr & = \frac{\TP}{\TP + \FP} \\[0.5em]
            \Re & = \frac{\TP}{\TP + \FN} \\[0.5em]
            F_1 & = \frac{2 \cdot \Pr \cdot \Re}{\Pr + \Re}
        \end{aligned}
    \end{equation}
\end{samepage}

\pagebreak[2]


\subsection{Evaluation Results}
\label{sec:evaluation_results}

The annotated data were split into a training, test, and validation set using a ratio of 70:20:10. Splitting was performed at segment level, not at token level, to avoid breaking multi-word entities.

\subsubsection{NER Accuracy}
\Cref{tab:ner} reports the results of the separate NER step, evaluated on the historical and contemporary real-world data.
The NER model contains multiple stochastic components.
To ensure statistical robustness, the model was trained five times using different random seeds and the standard deviation of the evaluation metrics was reported.

\pagebreak[1]

\begin{table}[t]
    \caption{Accuracy for Named Entity Recognition.}
    \label{tab:ner}
    \centering
    \begin{tabularx}{\columnwidth}{Lrrr}
        \toprule
        & \textbf{Precision} & \textbf{Recall} & \textbf{\FOne score} \\
        \midrule

        \begin{filecontents*}{data/ner.csv}
pr;re;fone
79.70 $\pm$ 2.41\,\%; 92.95 $\pm$ 1.11\,\%; 85.79 $\pm$ 1.33\,\%
\end{filecontents*}
\csvreader[
        head to column names,
        separator=semicolon,
        late after line=\\
        ]{data/ner.csv}
        {}
        {NER & \pr & \re & \fone}

        \bottomrule
    \end{tabularx}
    \vspace{0em plus 0.5em}
\end{table}



\pagebreak[2]

\subsubsection{Comparing Classifiers}
\label{sec:evaluation_results_classifiers}

\begin{table}[t]
    \caption{Accuracy of different classifiers.}
    \label{tab:classifiers}
    \centering
    \begin{tabularx}{\columnwidth}{Lr}
        \toprule
        \textbf{Classifier} & \textbf{\FOne score} \\
        \midrule

        \begin{filecontents*}{data/classifiers.csv}
classifier;fone
SVM;94.92\,\%
XGBoost;82.71\,\%
logistic regression;62.59\,\%
\end{filecontents*}
\csvreader[
        head to column names,
        separator=semicolon,
        late after line=\\
        ]{data/classifiers.csv}
        {}
        {\classifier & \fone}

        \bottomrule
    \end{tabularx}
    \vspace{0em plus 0.5em}
\end{table}

As mentioned in \Cref{sec:method_occupation_coding_classifiers}, three classifiers -- an SVM, XGBoost, and logistic regression -- have been systematically compared. \Cref{tab:classifiers} reports the \FOne scores of each of these classifiers, evaluated on a test set which was split from the standardized word lists, with the SVM achieving the highest accuracy (\FOne~score: 94.92\,\%).
Note that the accuracy on standardized word lists is significantly higher than when the model is later evaluated on real-world data, as reported in the following tables.

\pagebreak[3]

\subsubsection{One or Two Steps}
\label{sec:evaluation_results_one_two}

\begin{table}[t]
    \caption{Accuracy for one-step and two-step occupation coding.}
    \label{tab:accuracy}
    \centering
    \begin{tabularx}{\columnwidth}{Lrrr}
        \toprule
        \textbf{Approach} & \textbf{Precision} & \textbf{Recall} & \textbf{\FOne score} \\
        \midrule

        \begin{filecontents*}{data/accuracy.csv}
approach;pr;re;fone
one-step;38.23\,\%;60.78\,\%;46.94\,\%
two-step;54.16\,\%;60.75\,\%;57.26\,\%
\end{filecontents*}
\csvreader[
        head to column names,
        separator=semicolon,
        late after line=\\
        ]{data/accuracy.csv}
        {}
        {\approach & \pr & \re & \fone}

        \bottomrule
    \end{tabularx}
    \vspace{0em plus 0.5em}
\end{table}

\Cref{tab:accuracy} presents the central outcome of this study: A two-step occupation coding approach leads to substantially higher recognition rates than one-step approaches.

The model for the two-step approach has been trained using the same training data as the one-step approach implemented in the \emph{occupationCoding} R package, namely the list of occupations provided by the Federal Employment Agency. This demonstrates that a two-step approach offers conceptual advantages and yields higher recognition rates (\FOne score: 57.26\,\% compared to 46.94\,\%).
The improvement in \FOne score is mainly driven by a higher precision, while recall remains almost unchanged. This indicates that the advantage of the two-step approach lies in reducing false positives, as conceptually expected:
In the one-step setting, occupation coding is applied directly to free text and has to solve entity identification and classification simultaneously. As a result, non-occupational fragments, OCR noise, or context words may be incorrectly matched to occupations.
In the two-step setting, the NER model first restricts the input of the occupation classifier to occupational entities. This task separation reduces the number of irrelevant candidate strings and thereby improves precision. The number of false positives is considerably lower for the two-step approach.


\subsubsection{Confidence Thresholding}
\label{sec:evaluation_results_confidence}

\begin{table}[t]
    \caption{Confidence thresholding approaches.}
    \label{tab:confidence}
    \centering
    \begin{tabularx}{\columnwidth}{Lr}
        \toprule
        \textbf{Thresholding approach} & \textbf{\FOne score} \\
        \midrule

        \begin{filecontents*}{data/confidence.csv}
approach;fone
Baseline;34.07\,\%
Absolute threshold;43.89\,\%
Confidence margin;43.96\,\%
Combined (absolute \& margin);43.96\,\%
\end{filecontents*}
\csvreader[
        head to column names,
        separator=semicolon,
        late after line=\\
        ]{data/confidence.csv}
        {}
        {\approach & \fone}

        \bottomrule
    \end{tabularx}
    \vspace{0em plus 0.5em}
\end{table}

\Cref{tab:confidence} lists the \FOne scores for the baseline approach (\ie accepting all predictions without any threshold), the absolute threshold approach, the confidence margin approach, and the combination of both methods.

Both thresholding strategies outperform the baseline approach. Evaluated on historical and contemporary real-world data, the confidence margin approach achieves a slightly higher accuracy (\FOne score: 43.96\,\%) than the absolute threshold approach (\FOne score: 43.89\,\%).
The combination of both approaches, in this use case, achieves an identical \FOne score as the confidence margin approach.
The comparison is intended to illustrate the intuition behind both thresholding strategies and to introduce confidence-margin thresholding as an alternative to common absolute thresholding. Statistically robust conclusions would require a broader evaluation across multiple data splits and repeated runs.

\subsubsection{Benchmark Dataset}
\label{sec:evaluation_results_benchmark}

\newcounter{jobbertfn}
\setcounter{jobbertfn}{\value{footnote}+1}

\begin{table*}[t]
    \caption{Accuracy on the JobBERT dataset.}
    \label{tab:eval_jobbert_data}
    \centering
    \begin{tabularx}{\textwidth}{
        L@{\hspace{0.25em}}
        l@{\hspace{0.75em}}
        l@{\hspace{0.75em}}
        r@{\hspace{0.75em}}
        r
    }
    \toprule
    \textbf{Method} & \textbf{Concept} & \textbf{Lang.} & \textbf{Recall@1} & \textbf{\FOne score} \\[0.5em]
    \makecell[tl]{SBERT \\\citep{Reimers2019}\footnotemark[\value{jobbertfn}]} & \makecell[tl]{contextual\\information}  & EN & 19.30\,\% & ~\\[1.5em]
    \makecell[tl]{JobBERT \\\citep{Decorte2021}\footnotemark[\value{jobbertfn}]} & \makecell[tl]{contextual\\information}  & EN &  19.19\,\% & ~\\[1.5em]
    \makecell[tl]{Fine-tuned JobBERT \\\citep{Decorte2021}\footnotemark[\value{jobbertfn}]} & \makecell[tl]{contextual\\information}  & EN & 22.48\,\% & ~\\[1.5em]

    \begin{filecontents*}{data/eval_jobbert_data.csv}
approach;concept;fone
Proposed method (two-step, combined threshold);classification;44.20\,\%
\end{filecontents*}
\csvreader[
    head to column names,
    separator=semicolon,
    late after line=\\
    ]{data/eval_jobbert_data.csv}
    {}
    {\approach & \concept & DE & ~ & \fone}

    \bottomrule
    \end{tabularx}
    \vspace{0em plus 0.5em}
\end{table*}

\footnotetext[\value{jobbertfn}]{Evaluation results from \cite{Decorte2021}.}

In the field of occupation coding, there is no standard benchmark dataset.
However, \citet{Decorte2021} published a dataset\footnote{See \url{https://github.com/jensjorisdecorte/JobBERT-evaluation-dataset}.} to evaluate their JobBERT approach. This test set is a subset of the ESCO dataset, containing 15\,462 occupation instances and 1\,369 unique occupations.
We evaluated the proposed two-step approach on this JobBERT dataset.

It should be emphasized that the different approaches are only partially comparable, as they rely on different methodologies. JobBERT \cite{Decorte2021} and the Sentence-BERT (SBERT) approach \cite{Reimers2019} are based on \emph{contextual information} obtained through semantic analysis of skill-related data. In contrast, the proposed method is based on the recognition and classification of specific entities.
Conceptually, semantic-based approaches are superior to entity-based ones. However, they must be properly trained; otherwise, entity-based approaches can indeed outperform semantic-based ones.

\pagebreak[2]

To evaluate the proposed two-step approach, we performed a document-level evaluation on the original ESCO HTML pages\footnote{See \url{https://github.com/alexander-esser/jobbert_html_dataset}.}. The goal is to find the ground truth occupation with the corresponding job ID on the HTML page.
If multiple occupations were found, the most frequent job ID, weighted by the confidence, was selected as the final prediction for the page (weighted majority vote).
A prediction is counted as correct only if the occupation has been recognized during NER and the classification result is correct.
In cases where the classifier predicted multiple job IDs, all candidates were considered and contributed equally to the final decision. Samples without a valid ground truth job ID were excluded from the evaluation.

For confidence thresholding, a combination of an absolute threshold (empirically chosen margin: $t = 0.6$) and a confidence margin approach (empirically chosen margin: $m = \text{top}_1-\text{top}_2 = 0.3$) was applied.

The JobBERT test data are in English. In contrast, the proposed two-step approach has been trained on German data. However, since ESCO is multilingual, a mapping between English and German occupations could be established.
The two-step approach was evaluated on the same JobBERT data, only in German instead of English.
In this work, the focus was to demonstrate the fundamental conceptual advantages of a two-step approach. In future work, we plan to extend our approach by training multilingual models.

\Cref{tab:eval_jobbert_data} reports the accuracy of three established approaches, SBERT \citep{Reimers2019}, JobBERT, and a fine-tuned JobBERT variant \cite{Decorte2021}, as well as the proposed two-step approach. The evaluation results for the first three methods are taken from \citep{Decorte2021}. The reported \emph{micro Recall@1} (\ie, the top-ranked prediction) most closely corresponds to the \FOne score used for the two-step approach. Although the results are only partially comparable, they nonetheless indicate that the proposed method achieves at least a comparable level of accuracy, highlighting the potential of the approach.
Establishing a standardized benchmark dataset for occupation recognition and classification remains an avenue for future work.

For the proposed method, recall (82.86\%) is substantially higher than precision (30.14\%).
From a theoretical perspective, a high recall indicates that the system is highly sensitive and captures most job titles, resulting in few false negatives.
Such a coverage-oriented system identifies most of the potential job titles for subsequent analysis, albeit at the cost of reduced precision, including some incorrect predictions.

Regarding the effect of multi-stage fine-tuning and NAT on the recognition of vocational entities, a detailed analysis is provided in our prior work \citep{Esser2026, Esser2026a}.

\pagebreak[2]


\section{Conclusion}
\label{sec:conclusion}

This paper presented a two-step occupation coding approach that deliberately separates occupation coding into two distinct tasks: (i) detecting job titles in continuous text and (ii)~mapping the identified titles to an occupational taxonomy.

In the first \gls{NER} step, multi-stage fine-tuning was employed to incorporate additional training data and \gls{NAT} to improve the robustness against typical sources of noise, such as spelling errors.
In the second OC step, several classifiers were evaluated, with an SVM yielding the strongest performance. It was demonstrated that using a confidence margin, rather than a fixed absolute confidence threshold, leads to additional small improvements in recognition accuracy.

\subsection*{Limitations and Future Work}
For occupation coding, we used classical machine learning classifiers to maintain comparability with one-step approaches and focus on demonstrating the conceptual benefits of task separation.
For practical applications, training an advanced classifier, such as a Transformer-based model, promises to further enhance recognition rates.

Currently, the approach relies on explicitly mentioned job titles, which is error-prone, especially in contemporary social media data where occupations are frequently paraphrased rather than explicitly stated \cite{Esser2026b}.
Future work will, therefore, focus on recognizing occupations based on semantic context, in a way similar to JobBERT or SBERT, but as a two-step approach. The NER model introduced in \citep{Esser2026, Esser2026a} already supports the recognition of multiple entity types, such as skills, working activities, or work equipment. This provides a foundation for a more comprehensive semantic understanding and for further developing the method towards practical application.

From a practical perspective, the results should be interpreted as a methodological improvement rather than a ready-to-use occupation coding system. For deploying the application in practice, further improvements in contextual recognition and classification accuracy are required.

Another goal of future work is to extend the system to English and other languages. The KldB taxonomy is aligned with ISCO, making it possible to automatically generate annotated ground truth data and to train multilingual models.

In summary, this study demonstrated that the proposed two-step approach achieves substantial gains in accuracy, robustness, and interpretability compared to one-step occupation coding methods.

\pagebreak[3]


\needspace{8\baselineskip}
\section*{Acknowledgments}
{
\footnotesize

During the preparation of this manuscript, the authors used generative AI tools (ChatGPT, DeepL) for
spelling and grammar checks and wording suggestions.
The authors reviewed and edited all generated content and take full responsibility for the final text.

\smallskip
\noindent In addition, during the development of software components used in this study, generative AI tools (ChatGPT and GitHub Copilot) were used for assisted coding. All generated code was reviewed, tested, and integrated by the authors, who take full responsibility for the implemented software.
}

\pagebreak

\bibliographystyle{apalike}
\bibliography{references}

\end{document}